\documentclass{article}

\usepackage{microtype}
\usepackage{graphicx}
\usepackage{subcaption}
\usepackage{booktabs} 

\usepackage{hyperref}

\usepackage[accepted]{icml2026}

\usepackage{amsmath}
\usepackage{amssymb}
\usepackage{mathtools}
\usepackage{amsthm}
\usepackage{algorithmic}
\usepackage{algorithm}
\usepackage{multicol}
\usepackage{multirow}
\usepackage{makecell}
\usepackage{threeparttable}
\usepackage{array}
\usepackage{arydshln}
\usepackage{textcomp}
\usepackage{stfloats}
\usepackage{url}
\usepackage{verbatim}
\usepackage{graphicx}
\usepackage{bm}
\usepackage[table]{xcolor}

\usepackage[capitalize,noabbrev]{cleveref}

\theoremstyle{plain}

\theoremstyle{definition}

\theoremstyle{remark}

\usepackage[textsize=tiny]{todonotes}

\icmltitlerunning{FeDMRA: Federated Incremental Learning with Dynamic Memory Replay Allocation}

\begin{document}

\twocolumn[
  \icmltitle{FeDMRA: Federated Incremental Learning with \\
 Dynamic Memory Replay Allocation}



  \icmlsetsymbol{equal}{*}

  \begin{icmlauthorlist}
    \icmlauthor{Tiantian Wang}{hust}
    \icmlauthor{Xiang Xiang}{hust,uw}
    \icmlauthor{Simon S. Du}{uw}
  \end{icmlauthorlist}

  \icmlaffiliation{hust}{Huazhong University of Science and Technology, Wuhan, China.}
  \icmlaffiliation{uw}{University of Washington, Seattle, USA}

  \icmlcorrespondingauthor{Xiang Xiang}{xex@hust.edu.cn}

  \icmlkeywords{Machine Learning, ICML}

  \vskip 0.3in
]



\printAffiliationsAndNotice{}  

\begin{abstract}
  In federated healthcare systems, Federated Class-Incremental Learning (FCIL) has emerged as a key paradigm, enabling continuous adaptive model learning among distributed clients while safeguarding data privacy. However, in practical applications, data across agent nodes within the distributed framework often exhibits non-independent and identically distributed (non-IID) characteristics, rendering traditional continual learning methods inapplicable. To address these challenges, this paper covers more comprehensive incremental task scenarios and proposes a dynamic memory allocation strategy for exemplar storage based on the data replay mechanism. This strategy fully taps into the inherent potential of data heterogeneity, while taking into account the performance fairness of all participating clients, thereby establishing a balanced and adaptive solution to mitigate catastrophic forgetting. Unlike the fixed allocation of client exemplar memory, the proposed scheme emphasizes the rational allocation of limited storage resources among clients to improve model performance. Furthermore, extensive experiments are conducted on three medical image datasets, and the results demonstrate significant performance improvements compared to existing baseline models.
\end{abstract}
\begin{figure*}[htp]
	\centering
	\includegraphics[width=\textwidth]{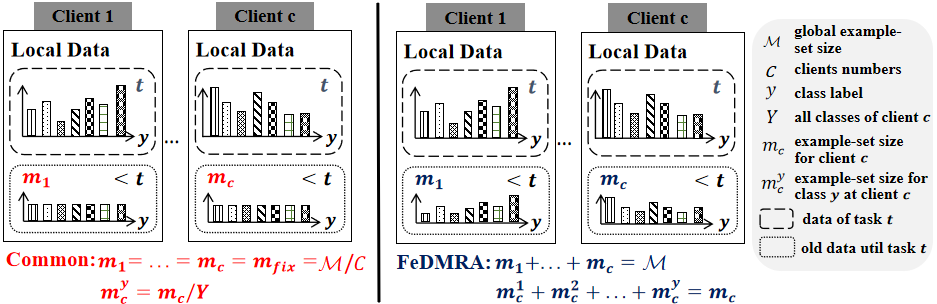} 
	\caption{The legend reflects our motivations and contributions. The coordinate system is used to characterize the sample distribution of the client. 
    Compared to \textbf{common} replay-based methods that allocate fixed memory size ${m_{fix}}$ per client and uniformly distribute it across all classes, \textbf{FeDMRA} leverages the heterogeneity of client data distributions and their performance contributions to dynamically allocate matched memory sizes ${m_{c}}$ and ${m_c^y}$, thereby optimizing the influence of stored exemplars.}
	\label{fig:instruction}
\end{figure*}
\section{Introduction}
With the gradual increase in the number of hematologic disease cases caused by living environments and lifestyle factors, accurate white blood cell (WBC) classification for blood disorder diagnosis has become a prominent research focus in recent years \cite{ZENG2023133865,BHATIA2023108913}. As deep learning technologies advance, researchers have proposed using computer vision techniques and data analysis to assist with WBC image classification tasks for hematological diagnosis \cite{wbc-method1,wbc-method2,2024Classification}. However, significant challenges persist during rare and complex case sampling, including collection difficulties, limited sample sizes, and noncentralized distribution patterns caused by objective factors such as population demographics, disease varieties, and medical equipment disparities. To address these issues, Federated Learning (FL) has been introduced as a distributed framework, allowing the training of collaborative global models among participants using localized datasets while ensuring the preservation of data privacy \cite{FL1, FL-Medical,li2020review, Wu2024FromOT, Wang2024AsymmetricML, guan2024federated,ng2021federated}.

 Conventional federated learning approaches typically assume static data distributions. However, in real-world healthcare scenarios, medical data from different clinical sites is usually generated in streaming or phased fashion, with both data distributions and categories evolving over time \cite{CL+MEDICAL,cl-me,feng2022clinical}, making these conventional approaches inadequate. To address this challenge, incremental learning has emerged as an effective solution. Nevertheless, a critical and unavoidable challenge in incremental learning is that models typically exhibit rapid and severe forgetting of previously acquired knowledge when learning new tasks or categories\cite{2017forget, van2024continual}.

 Recent years have witnessed numerous proposed solutions to address catastrophic forgetting in centralized learning paradigms \cite{icarl,ewc,Foster,lwf}. Among these, memory replay has been recognized as a straightforward and effective approach that has been extended to federated settings \cite{cha2020proxy,pennisi2024feder,refed,qi2023better}.In centralized environments, the incremental learning configuration resembles a single-client continual learning task, where replay-based methods typically employ predetermined and fixed exemplar memory allocations. In contrast, the federated setting involves multiple clients exhibiting non-independent and identically distributed (non-IID) data characteristics, with each client contributing unequally to the global model due to varying environmental conditions and resource constraints \cite{FLchallenge,ye2023heterogeneous,li2020federated,huang2022learn}.
This fundamental discrepancy raises significant doubts about the direct applicability of conventional replay-based approaches in distributed frameworks. 

Furthermore, most existing studies tend to develop solutions for isolated incremental learning scenarios. For example, studies \cite{mittal2021essentials,masana2022class,belouadah2019il2m,pian2023audio,Guo2024PILoRAPG,FCIL} address class-incremental learning, while \cite{xie2022general,wang2022s,wang2024non} focus on domain-incremental learning. However, real-world applications rarely conform to such idealized conditions, as the emergence of novel categories and distribution shifts typically occur concurrently in practice.

To address these challenges, we propose FeDMRA, \textit{federated incremental learning with dynamic memory replay allocation}: a novel exemplar memory allocation framework based on efficient data replay. Specifically, before the arrival of new tasks, the server dynamically allocates memory portions to each client based on two key factors: (1) the client's private data distribution, and (2) its contributions in global fusion. This innovative approach enables adaptive adjustment of the number of stored old samples per client during incremental training. A more intuitive explanation is given in Fig.~\ref{fig:instruction}. 

Our principal contributions can be summarized as follows: \\
$\bullet$ We analyze challenges in practical applications and construct a more comprehensive framework for Federated Continuous Learning (FCL), encompassing: Federated Class Incremental Learning (FCIL), Federated Domain Incremental Learning (FDIL), and Federated Class-Domain Incremental Learning (FCDIL).\\
$\bullet$ We analyzed the potential issues in existing continual learning schemes based on fixed exemplar set memory and designed a new federated continual learning approach. Furthermore, we focused on the memory allocation strategies for exemplar sets on each client to address the unfairness caused by data heterogeneity.\\
$\bullet$ We optimized client training under traditional federated learning techniques by integrating regularization and knowledge distillation methods to alleviate the problem of catastrophic forgetting.\\
$\bullet$ We demonstrated the effectiveness of our method on datasets that are clinically significant and challenging.

\section{Related Work}
Federated learning (FL), as a mainstream distributed learning framework, has emerged as a research hots pot in both academia and industry due to its core advantages of eliminating the need for raw data upload, while balancing privacy protection and efficient data utilization \cite{fedavg,zeng2024tackling,ren2024federated,hsu2020federated,fair-FL}. However, the streaming dynamic growth of client-side samples over time places higher demands on the generalization ability and sustainable learning performance of the model, thus giving rise to Federated Continual Learning (FCL). Catastrophic forgetting \cite{van2024continual} constitutes the core challenge in FCL, and the academic community has conducted extensive research and proposed various solutions to address this issue, including:
regularization-based approaches \cite{chen2022multi,mittal2021essentials,liu2022few}, replay-based methods, knowledge distillation techniques\cite{kang2022class,zhao2023few,cheraghian2021semantic,zhu2021data}, and data generation strategies \cite{MFCL,DDDR}. However, to the best of our knowledge, {most existing studies typically focus on isolated task scenarios}, such as CIL or DIL. Research on practical and complex CDIL tasks remains limited. This gap is particularly salient in federated healthcare applications, where edge data typically exhibits category expansion and distributional shifts across institutions. We explicitly tackles this underexplored yet critical setting, proposing a unified framework to handle above challenges in FCL.

FCL has been studied for categories, distribution domains, and tasks. 
Extending and deriving from centralized algorithms as a solution for FCL is straightforward and effective, but it does not conform to the actual federated settings. As described in \cite{Wang2024AsymmetricML, li2020federated, FCIL, zhao2023few}, class imbalance between old and new classes is a key challenge faced by exemplar replay methods. FCIL\cite{FCIL} designs a proxy server to select the best old global model to balance the bias. ReFed \cite{refed} proposes to train a private information model to select more effective old samples to resist forgetting. MFCL \cite{MFCL} and DDDR \cite{DDDR} generate samples that conform to the distribution of old data through generative models and add them to the training of new tasks. FedSpace \cite{FedSpace} proposes a method of asynchronous class-incremental learning based on pre-training and prototype enhancement. Generally speaking, all of them are based on setting {equal memory sizes for the exemplar sets of clients and categories}. However, this is not applicable to the current situation of unbalanced data and diverse resources at each edge side under the federated settings. Balancing the data distribution of clients and their performance contributions, and dynamically allocating the memory of the exemplar set for them are the issues that this paper focuses on.

\section{PRELIMINARIES \& PROBLEM DEFINITION}
This section lays out the foundational framework and core concepts that are essential for understanding the problem definition and proposed solutions that follow. Our investigation centers on the phenomenon of catastrophic forgetting within FCL models. The primary objective is to preserve the model's classification accuracy in the face of challenges, such as distributional shifts, introduced by new sequential data.
\subsection{Problem Definition - FL}
FL is a distributed machine learning paradigm that aims to collaboratively train a shared global model among multiple clients while preserving data privacy. Consider a federated learning system with $C$ clients, indexed by $c\in {1,2,...,C}$. Each client $c$ possesses a local private dataset, $D_c$
 , which consists of $n_c = |D_c|$ data samples. A key constraint is that the data from any client $D_c$ never leaves its local device and is not shared with other clients or a central server. These datasets distributed across clients are typically heterogeneous, i.e., Non-Independent and Identically Distributed (Non-IID).
The  local objective for each client k is to find a set of model parameters, w, that minimizes the loss function $\mathcal{F}_c(w)$ on its local dataset $D_c$. This local loss function is typically defined as the empirical risk:
\begin{equation}
    \mathcal{F}_c(w)=\frac{1}{n_c}\sum_{i\in\mathcal{D}_c}\ell(x_i,y_i;w)
\end{equation}
where $(x_i,y_i)$ is a data sample from $D_c$, and $\ell(\cdot)$ is the loss function for a single sample (e.g., cross-entropy loss). The global objective of FL is to minimize the weighted average of the local losses over all clients, denoted as the global loss function $\mathcal{F}(w)$. The weight for each client is typically the fraction of its data size relative to the total data size, $N=\sum_c^C n_c$.
Therefore, the central problem in Federated Learning can be formally defined as the following optimization problem:
\begin{equation}
    \min_w \mathcal{F}(w) \triangleq \sum_{c=1}^C\frac{n_c}{N}\mathcal{F}_c(w)
     \label{eq.FL}
\end{equation}
This optimization must be performed under several key constraints, including data privacy, statistical heterogeneity (Non-IID data), and limited communication bandwidth.
\subsection{Problem Definition - FCL}
As an extension of FL, FCL enables multiple clients to collaboratively train a global model under the condition of dynamically and sequentially arriving data.
In FCL, clients progressively update their models following task sequence stream $\{D^1,D^2,...,D^T \}$, where each element $ D^t$ represents a task-specific dataset. Formally, we define ${D^t=\{x_i^t, y_i^t\}_{i=1}^{N^t}}$ as $N^t$ paired samples comprising input data $x_i^t\in X^t$ and corresponding label $y_i^t\in Y^t$. $X=\cup_{t=1}^TX^t,Y=\cup_{t=1}^TY^t$  represent the sample domain space and the class space of all sequential tasks respectively. 

Therefore, the objective of FCL is for all participating parties to continually learn a shared model across a sequence of $T$ tasks, Eq.~\ref{eq.FL} can be reformulated as the optimization problem:
\begin{equation}
    w_{g,t}=\arg \min _{w} \sum_{c=1}^{C} \frac{1}{\bar{N}^{t}}\sum_{i \in \bar{D_c^t}} \ell (\bar{x}_{c,i}^t,\bar{y}_{c,i}^t;w)
\end{equation}
where $\bar{D_c^t}= D_{c}^t\cup D_{c,cache}^{t}$, $D_c^t$ represents the data sequence of client $c$ during task $t$, ${D_{c,cache}^{t}}$ is example set to cache old data until task $t$ has a size of $m_{c,t}$, and $\bar{N}_{c}^{t}=m_{c,t}+N_{c}^{t}$. 

In existing FCL research, the replay mechanism is considered the simplest yet most effective approach, which provides equivalent storage capacity for exemplar sets across all clients, i.e., ${m_{fix}}$ = ${m_{c,t}^y}$. In contrast, we propose a global exemplar memory pool that equals the union of all clients' local exemplar sets, with the expression:
\begin{equation}
    \mathcal{M}=\sum_{c \in C}\sum_{y \in Y}{m_{c,t}^y}, \, and \,\, m_{c,t}^y \in [m_{min},{m_{max}}]
\end{equation}
where $m_{max}$ denotes the maximum storage capacity supported by edge devices, and $m_{min}$ denotes the lower bound on the size of the exemplar set, preventing the failure to store old data due to extreme distributions.

\section{FeDMRA FRAMEWORK}
\begin{figure*}[t]
    \centering
    \includegraphics[width=0.9\textwidth]{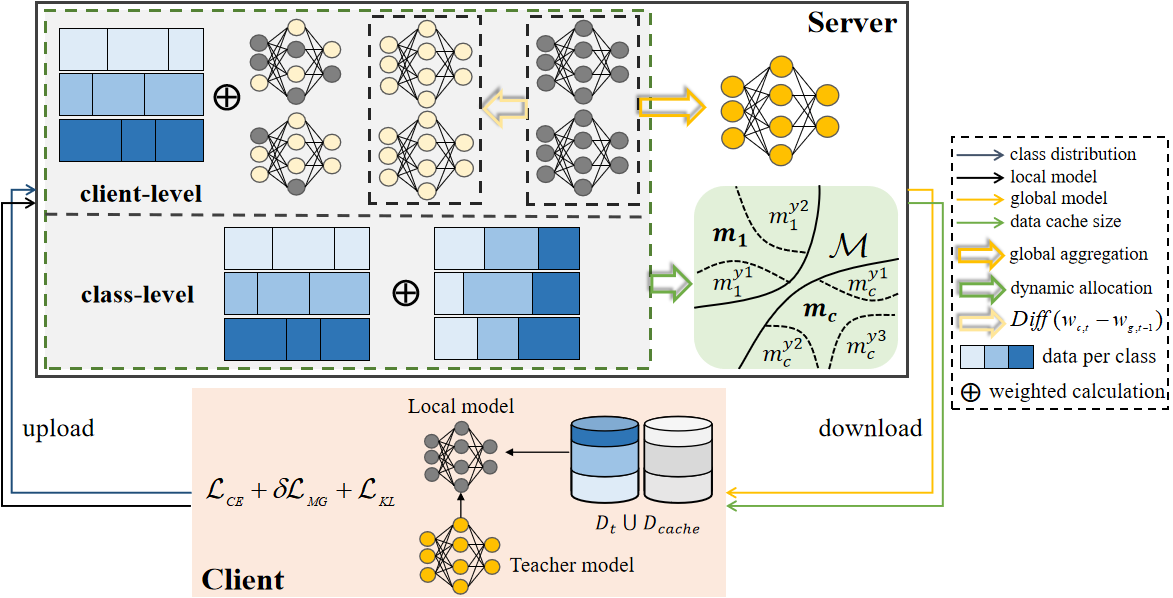}
    \caption{An overview of the proposed FeDMRA. \textbf{Server}:
    1. Receive the model parameters $w_{c}$ and class distribution uploaded by the clients. 2. Calculate the storage size of example sets for each category on the client side through client-level and class-level calculations. 3. Aggregate the client models using FedAvg to update the global model. \textbf{Client}: 1. Filter samples to fill the allocated size ${m_c^y}$ of the example set and replay. 2. Use the global model $w_{{g,t}}$ as a teacher model to perform knowledge transfer for the local model. 3. Local incremental training updates the model and uploads it.
    }
    \label{fig:main}
\end{figure*}

This paper proposes a novel framework called FeDMRA, which effectively mitigates catastrophic forgetting caused by data heterogeneity by dynamically allocating specific memory budgets to different clients' sample sets. The diagram provides a clearer illustration. A clearer illustration is provided in Fig.~\ref{fig:main}
\subsection{Server: Adaptive Example-set Memory Allocation}
\begin{algorithm}[t]
    \renewcommand{\algorithmicrequire}{\textbf{Input:}}
	\renewcommand{\algorithmicensure}{\textbf{Output:}}
	\caption{FeDMRA: Dynamic Allocation of Example Set Memory}
    \label{1}
    \begin{algorithmic}[1] 
        \REQUIRE  Number of clients: $C$, communication rounds: $R$, global memory cache size: $\mathcal{M}$, Initialized model parameters: $w_{g,0}$; 
	    \ENSURE the size of the example set for each class on the client side:
    $\{{m_{c,t}^y}, y \in Y^t,c\in C\}$; 
        \FORALL {$i=1,2,\cdots,T$}
            \STATE Received the local model parameters $w_{c,t}$ and class distributions uploaded from each client;
           
            \STATE ${b_{c,t}},{d_{c,t}}\leftarrow$ The model space contribution index and the sample space contribution index of the client are obtained through Eq.(5) and Eq.(6);\\
            \STATE $\{{m_1^t},{m_2^t},...,{m_c^t}\}\leftarrow$ Divide $\mathcal{M}$ according to Eq.(7);\\
               \FOR{$c=1,2,\cdots,C$}
                \STATE ${m_{c,t}^y}\leftarrow$ Calculate the memory of the example set for each class by Eq.(8); 
               \ENDFOR
            \STATE Perform global fusion and update to obtain 
             $w_{g,t}$.
           
        \ENDFOR
    \end{algorithmic}
\end{algorithm}
To mitigate the impact of data distribution on global performance, the server evaluates the contribution of client data to global aggregation updates through a dual-channel importance measurement mechanism, and dynamically allocates local exemplar memory for each client. This allocation process is divided into two phases: \textit{Client-level Allocation} and \textit{Class-level Allocation}. A detailed workflow summarized in Algorithm.~\ref{1}.

\textbf{Client-level Dynamic Allocation.} In the FCL framework, after each task is completed, the client sends its local updates (the model parameters $w_{c,t}$) to the central server. The server then performs an aggregation update to generate the next iteration of the global model. However, relying solely on parameter updates offers a limited perspective: while they reveal the trajectory of local optimization, they fail to capture the underlying data characteristics that drive such updates, potentially leading to a partial or incomplete understanding.

To address this limitation, we propose that clients concurrently transmit statistical metadata summarizing their local data distribution for the current task, such as the set of class labels and the corresponding number of samples per class. This approach provides the server with valuable data-centric insights without compromising the privacy of any raw local data. Subsequently, during the model aggregation phase, the server calculates a contribution index, denoted as $b_{c,t}$, for each local model. This index is determined by quantifying the deviation of the local model's update relative to the previous global model $w_{g,t-1}$:
\begin{equation}
    {b_{c,t}} = \frac{{Diff}({w}_{c,t}, {w}_{g,t - 1})}{\sum_{c\in C} {Diff}({w}_{c,t}, {w}_{g,t - 1})}
\end{equation}
where $Diff(\cdot)$ denotes the deviation between local and global models can be quantified either through vector subtraction $({w}_{c,t}-{w}_{g,t - 1})$. 
Furthermore, to compute the sample importance metric ${d_{c,t}}$, we conduct in-depth analysis of local data distributions that simultaneously considers local-level sample influence, and global-level significance across clients. 
This framework actively transforms heterogeneous distributions into driving forces for model enhancement:
\begin{equation}
    {d_{c,t}} = \frac{{\sum_y^{Y^t} {{N_{c,y}^t}}/{{N_{y}^t}}}}{\sum_c^C\sum_y^{Y^t}{{N_{c,y}^t}}/{{N_{y}^t}}}
\label{client m}     
\end{equation}
where ${N_{c,y}^t}/{{N_{y}^t}}$ represents the proportion of class $y$ in client $c$ relative to the global distribution of class $y$.

In summary, the server performs weighted aggregation of dual-channel importance indices to determine final client significance, subsequently allocating global example memory proportionally:
\begin{equation}
    {m_{c,t}} = \frac{d_{c,t}+b_{c,t}}{\sum_c^C({d_{c,t}+b_{c,t}})} \times \mathcal{M}
\end{equation}

\textbf{Class-level Dynamic Allocation.} Given the inherent heterogeneity of federated clients, after computing client-level memory allocations at the server, we further investigate the local and global distributions of each class to determine category-specific memory assignments. The naive approach of allocating equal storage space to both classes (i.e., $m_{c,A}=m_{c,B}$) is insufficient for adequately preserving samples of Disease A, which ultimately degrades the model's identification performance for this critical category. Therefore, by incorporating both intra-client and inter-client class distributions, the class-specific replay storage space is:
\begin{equation}
m_{c,t}^{y} = Norm\Biggl(
(1-a)\frac{N_{c,y}^{t}}{N_{c}^{t}} + a\frac{N_{c,y}^{t}}{N_{y}^{t}}
\Biggr) \times m_{c,t}
\label{client m}
\end{equation}
${m_{c,t}^y}$ denotes the allocated storage memory for class $y$ at the client $c$. $a$ is the hyperparameter that balances the proportion of the local and global distributions. $N_{c,y}^t$ is the number of samples of class $y$ for client $c$. $Norm(\cdot)$ represents the normalization operation to ensure that the sum of the number of examples across all classes is equal to {$m_{c,t}$}. 

\textbf{Global Aggregation.} During the server aggregation phase, the global model parameters $w_{g,t}$ are obtained through weighted aggregation following the classic FedAvg protocol\cite{fedavg}.

\subsection{Client: Continual Learning}
\label{client continual learning}
With the arrival of new tasks, each client trains and optimizes its model on the new data sequence. To mitigate forgetting, we construct a sample set containing representative old samples and integrate it into the training process for new tasks. Simultaneously, considering the heterogeneity of data across clients, we design an augmented objective function that minimizes loss while constraining the optimization direction, thereby reducing conflicting updates among clients.

\textbf{Example Set Selection.} Following the exemplar selection strategy proposed in \cite{refed}, this approach addresses the shortcomings of conventional methods that struggle to identify samples contributing beneficially to global updates. Inspired by momentum optimization techniques, the strategy maintains a client-specific personal information model on each client. This model integrates the historical global model $w_{g,t-1}$ and undergoes iterative optimization using data from the first $t-1$ tasks. The $t$-th new task arrives, the personal information model updated by:
\begin{equation}
\begin{aligned}v_{c,e}^{t}=&v_{c,e-1}^{t-1}-\eta\left(\sum_{i}\nabla \mathcal{L}\left(\mathcal{F}_{v_{c,e-1}^{t-1}}\left(\bar{x}_{c,t-1}^{i},\bar{y}_{c,t-1}^{i}\right)\right)\right) \\& +q(\lambda)\left(v_{c,e-1}^{t-1}-w_{g,t-1}\right)
\end{aligned}
\label{momentum-based}
\end{equation}
 where $q(\lambda)=\frac{1-\lambda}{2\lambda},\lambda\in(0,1)$, $\eta$ is the step of update and $e$ is the number of iterations. During the iterative optimization of the aforementioned objective function, samples with larger gradient norms are stored in the exemplar set.

In contrast to \cite{refed} where the size of $D_{c,cache}^{t}$ remains constant for all clients under each task, our work dynamically determines the size $m_{c,t}$ of $D_{c,cache}^{t}$ through server-side computation.
When constructing the example set $D_{c,cache}^{t}$ until task $t$
via Eq.\ref{momentum-based}, we select samples per class by sorting them in descending order of gradient norms. Based on Eq.\ref{client m}, the exemplar cache capacity for each category is computed, and the client selects $m_{c,t}^y$ exemplar samples for each class:
\begin{equation}
\begin{aligned}
D_{c,\text{cache}}^{t}
= \text{where } x_i \in
\bigl(
\underbrace{{
  \underbrace{\bar{G}_{x_1},\bar{G}_{x_2},\dots,}_{m_{c,t}^{y_1}}
  \bar{G}_{x_{N_{y_1}^{t}}}
}}_{N_{y_1}^{t}},
\dots,\\[4pt]
\underbrace{{
  \underbrace{\bar{G}_{x_1},\bar{G}_{x_2},\dots,}_{m_{c,t}^{y_2}}
  \bar{G}_{x_{N_{y_2}^{t}}}
}}_{N_{y_2}^{t}},
\dots,
\underbrace{{
  \underbrace{\bar{G}_{x_1},\bar{G}_{x_2},\dots,}_{m_{c,t}^{y_n}}
  \bar{G}_{x_{N_{Y}^{t}}}
}}_{N_{y_n}^{t}}
\bigr)
\end{aligned}
\end{equation}
where $\bar{G}_{x_i}=\sum G_{x_i}/e$ is the average gradient norm of sample $x_i$ over $e$ iterations, $G_{x_{i}}=
\left\|\nabla \mathcal{L}\left(\mathcal{F}_{v_{c,e-1}}^{t-1}\left(\bar{x}_{c,t-1}^{i},\bar{y}_{c,t-1}^{i}\right)\right)\right\|^2$ is the norm of the sample gradient. The algorithm flow is shown in Algorithm~\ref{2} in the appendix.

\begin{table*}[htp]
    \centering
    \caption{Report the classification accuracy $A_{last}$ and $A_{avg}$ under heterogeneity $\alpha=1.0$ (\%).}
\begin{tabular*}{\textwidth}{@{\extracolsep{\fill}}
c@{\hspace{1pt}}c@{\hspace{2pt}}c@{\hspace{2pt}}c@{\hspace{2pt}}c@{\hspace{2pt}}c@{\hspace{2pt}}c@{\hspace{2pt}}c@{\hspace{2pt}}c@{\hspace{2pt}}c@{\hspace{2pt}}c@{}}
        \toprule[1.2pt]
        \multirow{2}{*}{\textbf{Method}} & \multicolumn{2}{c}{\textbf{FCIL-M}} & \multicolumn{2}{c}{\textbf{FCIL-A}}& \multicolumn{2}{c}{\textbf{FCIL-H}}& \multicolumn{2}{c}{\textbf{FDIL}}&\multicolumn{2}{c}{\textbf{FCDIL}}\\
       
         & $\mathcal{A}_{avg}\uparrow$ & $\mathcal{A}_{last}\uparrow$& $\mathcal{A}_{avg}\uparrow$& $\mathcal{A}_{last}\uparrow$& $\mathcal{A}_{avg}\uparrow$& $\mathcal{A}_{last}\uparrow$& $\mathcal{A}_{avg}\uparrow$& $\mathcal{A}_{last}\uparrow$& $\mathcal{A}_{avg}\uparrow$& $\mathcal{A}_{last}\uparrow$\\
        \midrule
        \textbf{UP}&86.61&85.5&88.62&88.78&95.06&95.87&84.10&89.18&87.30&88.45\\
        \textbf{iCaRL+FL}& 72.05&58.88&65.36& 57.3&61.89 & 46.51&
         64.57&70.96&55.41 &55.56\\
        \textbf{UACL+FL}&52.06 &25.20&54.62 &49.32&66.59 & 48.44&51.21 & 36.53&51.93 &52.52\\
        \cdashline{1-11}
        \textbf{MFCL}&69.75&20.87&46.00 &31.91& 79.27 & 51.09& 74.02&64.82&51.12&29.16\\
        \textbf{DDDR}&68.14&64.34&51.38 &30.42& 78.23 & 64.00&77.56 &69.05&60.34&36.50\\
        \textbf{Re-Fed}&82.75&70.53&74.18 &70.84& 76.98 & 73.44&81.66 &87.48&80.74&67.72\\
        \cdashline{1-11}
        \rowcolor{lightgray!40}
        \textbf{FeDMRA}&83.06&80.63 &83.26&86.52& 91.77&87.34&82.71 & 87.64& 85.26& 83.35\\
        \bottomrule[1.2pt]
\end{tabular*}
    \label{tab:IL task}
\end{table*}

\begin{table*}[htp]
    \centering
    \caption{Report the classification accuracy $A_{last}$ and $A_{avg}$ under heterogeneity $\alpha=0.5$ (\%).}
\begin{tabular*}{\textwidth}{@{\extracolsep{\fill}}
c@{\hspace{1pt}}c@{\hspace{2pt}}c@{\hspace{2pt}}c@{\hspace{2pt}}c@{\hspace{2pt}}c@{\hspace{2pt}}c@{\hspace{2pt}}c@{\hspace{2pt}}c@{\hspace{2pt}}c@{\hspace{2pt}}c@{}}
        \toprule[1.2pt]
        \multirow{2}{*}{\textbf{Method}} & \multicolumn{2}{c}{\textbf{FCIL-M}} & \multicolumn{2}{c}{\textbf{FCIL-A}}& \multicolumn{2}{c}{\textbf{FCIL-H}}& \multicolumn{2}{c}{\textbf{FDIL}}&\multicolumn{2}{c}{\textbf{FCDIL}}\\
        
         & $\mathcal{A}_{avg} \uparrow$& $\mathcal{A}_{last}\uparrow$& $\mathcal{A}_{avg}\uparrow$& $\mathcal{A}_{last}\uparrow$& $\mathcal{A}_{avg}\uparrow$& $\mathcal{A}_{last}\uparrow$& $\mathcal{A}_{avg}\uparrow$& $\mathcal{A}_{last}\uparrow$& $\mathcal{A}_{avg}\uparrow$& $\mathcal{A}_{last}\uparrow$\\
        \midrule
        \textbf{UP}&86.61&85.50&88.62&88.78&95.06&95.87&84.10&89.18&83.61 &86.95 \\
        \textbf{iCaRL+FL}& 72.05&58.88&65.36& 57.30&61.89 & 46.51&
         64.57&70.96& 80.53&77.49\\
        \textbf{UACL+FL}&52.06 &25.20&54.62 &49.32&66.59 & 48.44&51.21 & 46.05&51.60 &53.63\\
        \cdashline{1-11}
        \textbf{MFCL}&56.40 &15.34 &39.25 &8.37& 77.91 & 51.20& 75.50&68.48 &38.92 &18.86\\
        \textbf{DDDR}&40.75&17.80&48.74 &28.15& 73.21 &59.02 &76.01 &72.30&41.17&14.72\\
        \textbf{Re-Fed}&82.75&70.53&74.18 &70.84& 76.98 & 73.44&73.85 &79.27&78.41 &63.45\\
        \cdashline{1-11}
        \textbf{FeDMRA}&77.29&70.25 &80.00&74.66& 88.48&82.95&75.83 & 79.02& 83.57& 85.79\\
        
        \bottomrule[1.2pt]
\end{tabular*}
    \label{tab:IL task0.5}
\end{table*}
\textbf{Local Updating.} For client-side training, we propose an optimization objective function. Firstly, as with most image classification tasks, we employ cross-entropy loss $\mathcal{L}_{CE}$ to minimize the divergence between the model's predicted distribution and the true distribution, thereby learning discriminative class-specific features.

Furthermore, as analyzed in \cite{celoss}\cite{zeng2024tackling}, the $\mathcal{L}_{CE}$ loss prompts the model to quickly learn more easily classifiable features by imposing heavier penalties on misclassifications. However, this approach may lead to the neglect of stable feature learning and is also one of the primary causes of catastrophic forgetting in incremental learning scenarios. Inspired by this, we introduce the $\mathcal{L}_{MG}$ loss, which minimizes the L2 norm of the model's output space to constrain the magnitude of model outputs. This promotes more stable training and enhances generalization capability via:
\begin{equation}
    \mathcal{L}_{MG}=\log (1 + \| \mathcal{F}_{w_{c,t}} (\bar{x}_{c,t}) \|_2^2)
\end{equation}

To prevent local models from overfitting to local data during training and thereby causing global objective drift, this paper introduces a regularization loss term, denoted as $\mathcal{L}_{KL}$, which enforces consistency by measuring the Kullback-Leibler (KL) \cite{KL} divergence between the output distributions of the local and global models. As formulated in Eq.~\ref{eq.kl}, minimizing this term achieves a dual objective: (1) it constrains local updates to prevent significant deviation from the global knowledge, thereby alleviating catastrophic forgetting; and (2) it preserves model plasticity, ensuring the model can effectively adapt to the novel information presented by the new task:
\begin{equation}
\mathcal{L}_{{KL}} = {KL} \left\{S\left(\mathcal{F}\left(\bar{x}_{c,t}; {w}_{g,t-1}\right)\right) \,\middle\|\,S\left(\mathcal{F}\left(\bar{x}_{c,t}; {w}_{c,t}\right)\right)\right\}
\label{eq.kl}
\end{equation}
and $S(\cdot)$ is softmax function.

In summary, the client's final objective function is:
\begin{equation}
    \mathcal{L}_{FINAL} = \mathcal{L}_{{CE}} +\mathcal{L}_{KL}+ \delta \mathcal{L}_{{MG}}
\end{equation}
where $\delta$ acts as a balancing factor to control the strength of the regular term.
\begin{table*}[htp]
\centering
    \caption{Average accuracy with different configurations under various heterogeneities $\alpha$ (\%).}
\begin{tabular*}{\textwidth}{@{\extracolsep{\fill}}
c@{\hspace{1pt}}c@{\hspace{1.5pt}}c@{\hspace{1.5pt}}c@{\hspace{1.5pt}}c@{\hspace{1.5pt}}c@{\hspace{1.5pt}}c@{\hspace{1.5pt}}c@{\hspace{1.5pt}}c@{\hspace{1.5pt}}c@{\hspace{1.5pt}}c@{\hspace{1.5pt}}c@{\hspace{1.5pt}}c@{}}
        \toprule[1.2pt]
        $\bm{\alpha}$ & \multicolumn{4}{c}{$0.5$} & \multicolumn{4}{c}{$1.0$}& \multicolumn{4}{c}{$5.0$}\\
        \cdashline{2-13}
        $\bm{a}$ & \multicolumn{2}{c}{$0.8$}& \multicolumn{2}{c}{$0.4$}
         & \multicolumn{2}{c}{$0.8$}& \multicolumn{2}{c}{$0.4$}& \multicolumn{2}{c}{$0.8$}& \multicolumn{2}{c}{$0.4$}\\ 
          \cdashline{2-3}\cdashline{4-5}\cdashline{6-7}\cdashline{8-9}\cdashline{10-11}\cdashline{12-13}
        $\bm{\lambda}$ & $0.8$&$0.4$&$0.8$&$0.4$ &$0.8$&$0.4$&$0.8$&$0.4$&$0.8$&$0.4$&$0.8$&$0.4$\\
        \midrule
        \textbf{FCIL-M} &76.19&77.08&76.91&\textbf{77.29}&82.12&82.58&82.44&\textbf{83.06}&80.97&80.87&82.66&\textbf{82.90}\\
        \textbf{FCIL-A} &78.45&79.36&\textbf{80.00}&78.36&82.75&82.27&82.91&\textbf{83.26}&\textbf{84.23}&83.80&83.77&83.89\\
        \textbf{FCIL-H} &87.83&\textbf{88.48}&88.26&88.07&91.51&91.16&\textbf{91.77}&91.72&91.71&\textbf{93.19}&91.03&91.75\\
\textbf{FDIL}&72.44&73.13&\textbf{75.83}&72.41&80.70&80.72&\textbf{82.71}&81.55&80.75&81.05&\textbf{81.97}&81.36\\
\textbf{FCDIL}&82.64&80.26&\textbf{83.57}&82.18&83.95&84.42&84.49&\textbf{85.26}&84.37&83.65&\textbf{85.02}&83.46\\
        \bottomrule[1.2pt]
\end{tabular*}
    \label{tab:data-heter}
  
\end{table*}
\section{EXPERIMENTS AND RESULTS ANALYSIS}
\subsection{Experiment Setup}
\textbf{Datasets and FCL Settings.} We selected three white blood cell datasets in the medical field for the experiment, including Matek-19 \cite{Matek-19}, Acevedo-20 \cite{Acevedo-20}, and \cite{dataset}. We adopt a similar setup to the literature \cite{UACL}, partitioning the three datasets as follows: \textbf{FCIL}: The acronyms FCIL-M, FCIL-A and FCIL-H are used to denote CIL on Matek-19, Acevedo-20 and \cite{dataset}, respectively;
\textbf{FDIL}: Each incremental stage consists of a dataset reflecting the shift in the distribution domain;
\textbf{FCDIL}: In more complex scenarios, the distribution of streaming data exhibits a combination of characteristics where both the label space and the domain space increase progressively. The dataset and task specifications are provided in Appendix~\ref{app:dataset and task}.

\textbf{Baselines.} To ensure fair comparison with other significant works, we follow the same protocol proposed in \cite{icarl} for constructing the FCL task. We evaluate all methods using the representative federated learning model FedAvg: \textbf{iCaRL\cite{icarl}+FL} and \textbf{UACL\cite{UACL}+FL}, three models specifically designed for federated incremental learning: \textbf{MFCL}\cite{MFCL}, \textbf{DDDR}\cite{DDDR} and \textbf{ReFed}\cite{refed}, and the upper performance bound \textbf{UP} under ideal conditions. We
report the final accuracy $A_{last}$ upon completion of the last
streaming task and the average accuracy $A_{avg}$ across all tasks.

\textbf{Implementation Details.} The implementation of this paper is all carried out on the PyTorch framework. All models are trained on NVIDIA RTX 4090 GPU with 24GB memory capacity and use ResNet18 as the backbone. FeDMRA is trained using the Adam optimizer and set the learning rate to 0.003. The input size is resized to 224 × 224 pixels, and the batch size is set to 64. In addition, the parameter settings for the global example set $\mathcal{M}$ and the maximum storable example set ${m_{max}}$ on the client side are different under different tasks. For FCIL on three datasets: $\mathcal{M}$=1200, ${m_{max}}$=400. For FDIL and FCDIL: $\mathcal{M}$=3000, and ${m_{max}}$=800. 

\subsection{Comparison Analysis}
\textbf{Test Accuracy.} We evaluate our approach under FCIL, FDIL, and FCDIL scenarios. Client data heterogeneity is simulated using a Dirichlet distribution Dir($\alpha$). Tab.~\ref{tab:IL task} compares the performance of various baseline algorithms when heterogeneity parameter $\alpha=1.0$. The iCaRL algorithm, which reuses original samples, outperforms generative methods like MFCL and DDDR that rely on pseudo-sample generation. This stems from generative models losing fidelity and introducing synthetic noise during sample reconstruction. An exception occurs in the high-resolution
HLwbc dataset, where its richer feature space enables generative methods to better preserve discriminative features. FeDMRA demonstrates significant advantages in FDIL and FCDIL tasks due to its dynamic memory allocation mechanism, whereas generative model-based approaches struggle to capture dynamic patterns in changing data distributions. Furthermore, the Re-Fed method performs poorly in most tasks, which we attribute to its combination of heterogeneous perception paradigm selection and direct storage of representative samples. We also conducted experiments with a more heterogeneous setting $\alpha=0.5$ in Tab.~\ref{tab:IL task0.5}. Although the performance trends align with Tab.~\ref{tab:IL task}, all methods exhibit significant degradation. This indicates that the long-tail distribution induced by extreme data heterogeneity inevitably negatively impacts model performance.

\textbf{Hyper-parameter.} In addition, we conduct experiments under different degrees of heterogeneity and further explore the settings of the parameters $a$ and $\lambda$.  
Tab.~\ref{tab:data-heter} presents the average performance of FeDMRA with varying configurations of key parameters. $a$ is the weight assigned to the global proportion when computing the data space importance for a given class. We find that $a=0.4$ is optimal in most settings, suggesting a relative robustness to data heterogeneity. $\lambda$, serves as the weight for the regularization term in the update of the local information model. The higher the value of $\lambda$, the heavier the penalty for iterative updates that deviate from the global model.
The results in the table confirm that when heterogeneity is strong, placing greater emphasis on the global distribution is beneficial; when heterogeneity is moderate, prioritizing the local data distribution yields better performance; and as the data approaches IID settings, the optimal strategy balances attention between global and local distributions.

\begin{figure*}[tp]
\centering
 \begin{minipage}{\linewidth}
        \centering
        \includegraphics[width=0.5\linewidth]{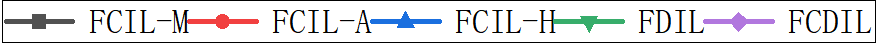}
    \end{minipage}
\subfloat[$\delta$]{\label{fig:subfig1}\includegraphics[width=0.325\textwidth]{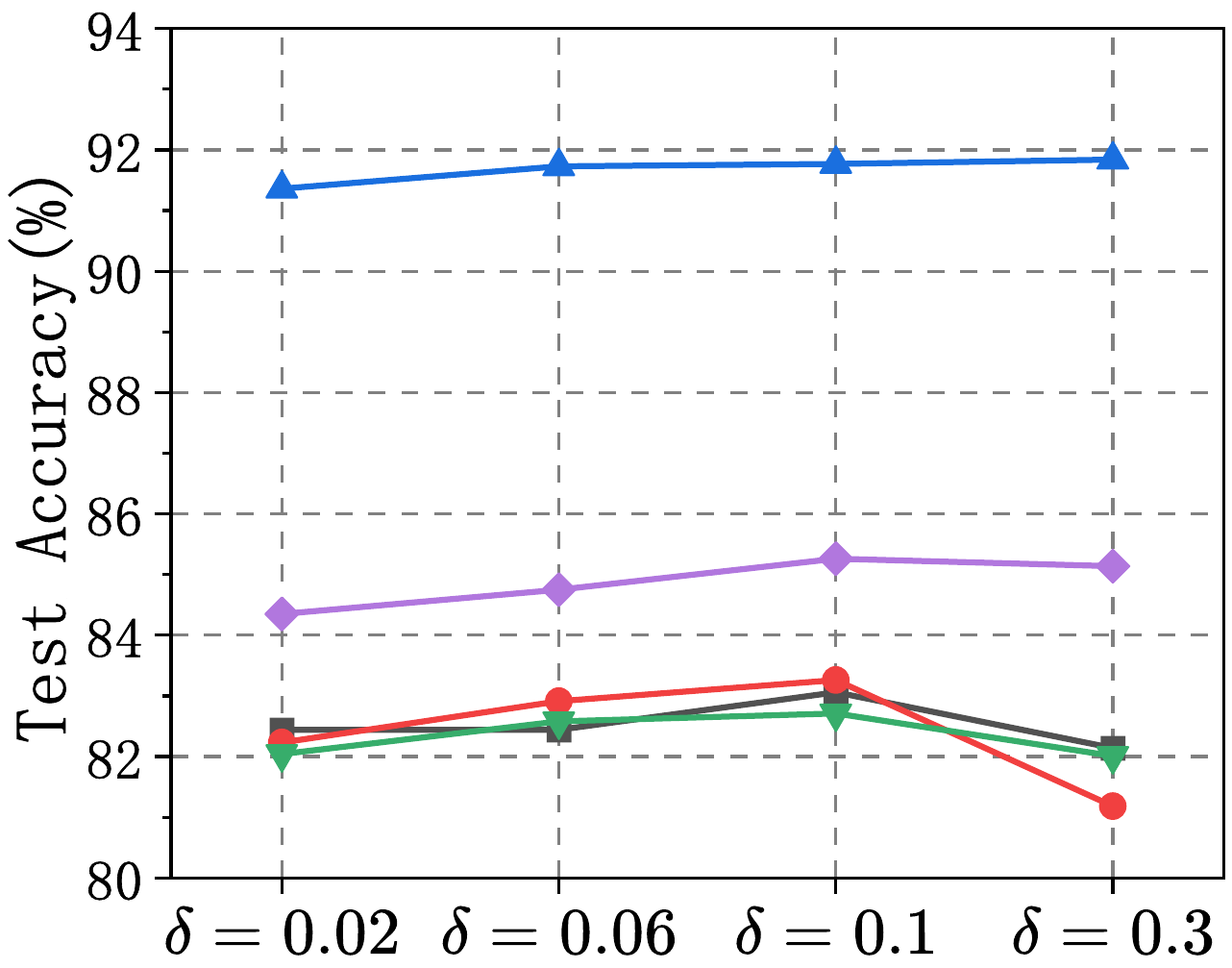}}
\subfloat[$m_{max}$]{\label{fig:subfig2}\includegraphics[width=0.325\textwidth]{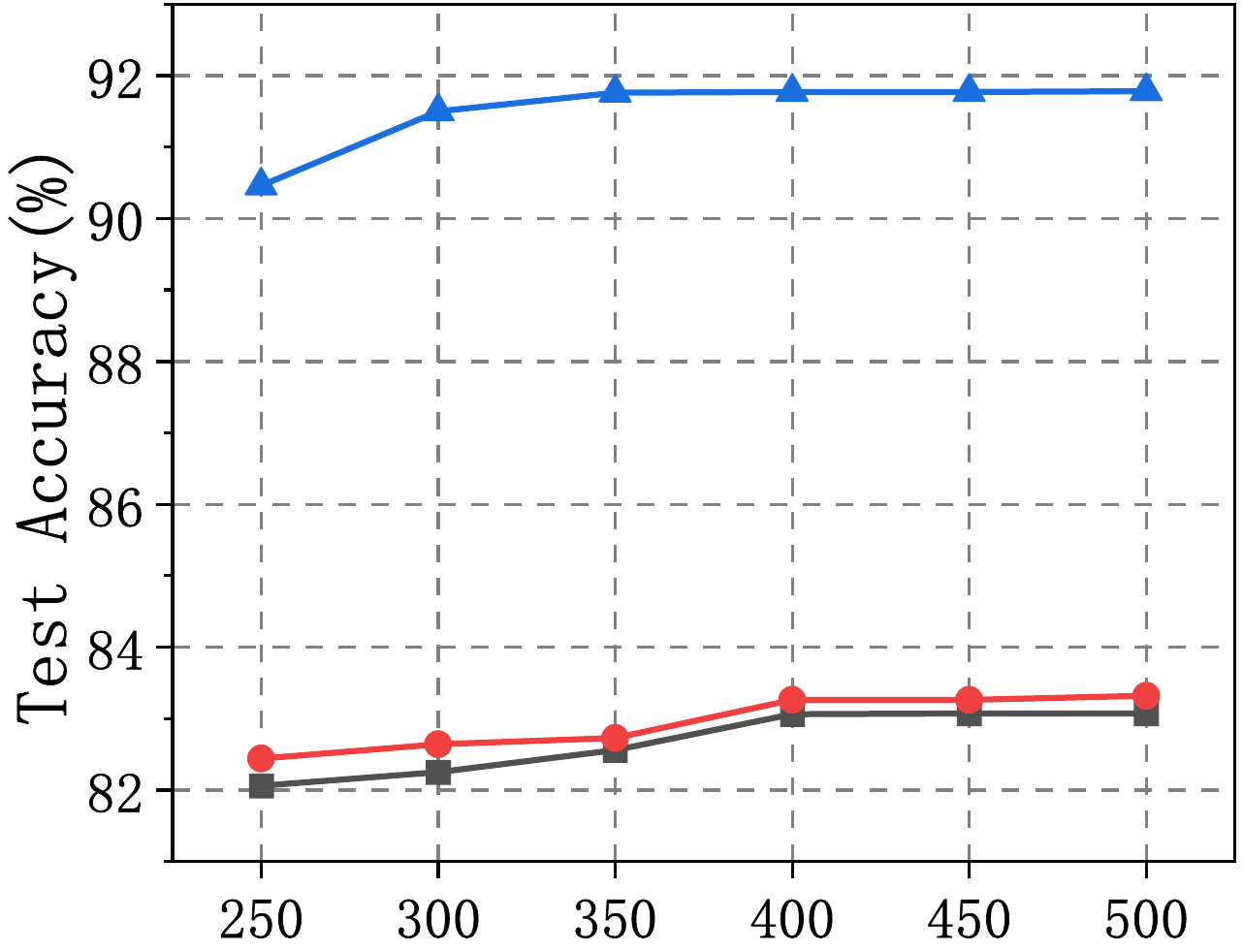}}
\subfloat[$m_{max}$]{\label{fig:subfig3}\includegraphics[width=0.325\textwidth]{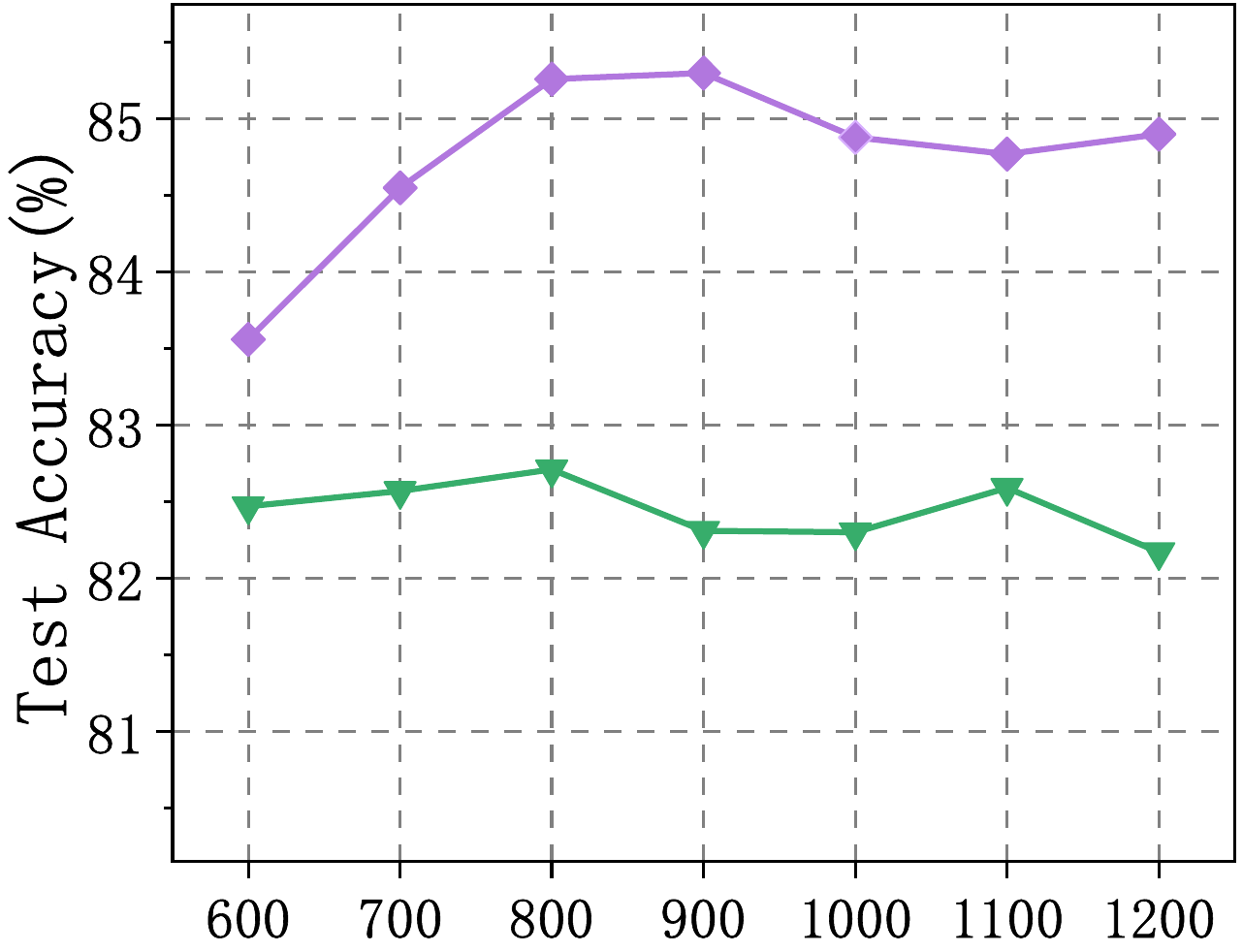}}
\caption{Our method under different configuration (a) weight $\delta$ of $\mathcal{L}_{MG}$, (b) $m_{max}$ for FCIL, (c) $m_{max}$ for FDIL and FCDIL.}
\label{fig:mainfig}
\vspace{-0.2cm}
\end{figure*}

\textbf{Important parameter settings.} We show the impact of other important parameters on the method in Fig.~\ref{fig:mainfig}. Specifically, (a) shows the the $A_{avg}$ with different weights $\delta$ of the $\mathcal{L}_{MG}$ loss. It can be seen that in most tasks, the optimal classification accuracy is achieved when $\delta=0.1$. (b) and (c) illustrate the impact of the maximum memory for storing example sets in clients under different tasks on performance. The sizes of the training sets on clients vary significantly across different tasks. Therefore, We configure varying global exemplar memory budgets and client-level maximum exemplar storage limits. For FCIL, $\mathcal{M}$=1200, and $m_{max}$ takes values in the range from 250 to 500 with an increment of 50. For FDIL and FCDIL, $\mathcal{M}$=3000, and $m_{max}$ takes values in the range from 600 to 1200 with an increment of 100. It is observed that for FCIL, the method tends to be stable when $m_{max}=400$. For FDIL and FCDIL, the algorithm achieves an ideal result when $m_{max}=800$. This experiment supports the results in Tab.~\ref{tab:IL task}.  

\begin{table*}[htbp]
    \centering
    \caption{Ablation study of FeDMRA (\%).}

\begin{tabular*}{\textwidth}{@{\extracolsep{\fill}}
c@{\hspace{1.5pt}}c@{\hspace{1.5pt}}c@{\hspace{1.5pt}}c@{\hspace{1.5pt}}c@{\hspace{1.5pt}}c@{\hspace{1.5pt}}c@{\hspace{1.5pt}}c@{\hspace{1.5pt}}c@{\hspace{1.5pt}}c@{\hspace{1.5pt}}c@{\hspace{1.5pt}}c@{\hspace{1.5pt}}c@{}}

 \toprule[1.2pt]
        {} & \multicolumn{4}{c}{\textbf{FCIL-H}}& \multicolumn{4}{c}{\textbf{FDIL}}&\multicolumn{4}{c}{\textbf{FCDIL}}\\
        \midrule
         $dy$&$\checkmark$& $\checkmark$ &$\times$&$\times$
         &$\checkmark$& $\checkmark$ &$\times$&$\times$
         &$\checkmark$& $\checkmark$ &$\times$&$\times$\\
       
         $\mathcal{L}_{MG}$&$\checkmark$& $\times$ &$\checkmark$&$\times$
         &$\checkmark$& $\times$ &$\checkmark$&$\times$
         &$\checkmark$& $\times$ &$\checkmark$&$\times$\\
        
         $\mathcal{L}_{KL}$&$\checkmark$& $\times$ &$\times$&$\checkmark$
         &$\checkmark$& $\times$ &$\times$&$\checkmark$
         &$\checkmark$& $\times$ &$\times$&$\checkmark$\\
         \cdashline{1-13}
         $A_{avg}\uparrow$ 
         &91.77&91.06&90.98&90.51
         &82.71&81.60&78.61&81.82
         &85.26&82.74&78.98&80.98\\
         $A_{last}\uparrow$ 
         &87.34&87.58&84.30&85.90
         &87.64&87.70&86.51&88.20
         &83.35&82.43&61.41&71.52\\
         \toprule[1.2pt]
\end{tabular*}
 \begin{tablenotes}
    \item[] * dy - Dynamic Memory Replay strategy
    \end{tablenotes}
    \label{tab:ablation1}
    \vspace{-0.1cm}
\end{table*}

\textbf{Ablation Study.} To verify the effectiveness of each component, we conducted ablation experiments on three datasets for different tasks in Tab.~\ref{tab:ablation1}. The results show that under most incremental tasks, the performance using only $dy$ can reach suboptimal levels. This conclusively demonstrates that leveraging client data distribution characteristics, overcoming fixed-memory constraints, and rationalizing exemplar storage allocation can effectively enhance network performance. Meanwhile, for $A_{avg}$, the $\mathcal{L}_{MG}$ component performs better, and for $A_{last}$, $\mathcal{L}_{KL}$ performs better, which also reflects the advantages of combining the two in terms of improving average performance and combating forgetting.

\section{Conclusion}
This paper extends to broader continual learning scenarios and proposes a novel federated incremental learning framework called FeDMRA. The framework introduces a dual-space dynamic evaluation mechanism for the central server, which combines the class distribution characteristics in the data sequences with the update shifts of model parameter sequences to adaptively allocate memory space for the example set, thus effectively overcoming the limitations of fixed memory budgets for each client in heterogeneous scenarios. At the same time, it designs optimization objectives to update the edge clients. This study is not constrained by the data storage format and provides a new perspective for replay-based methods. Extensive experiments validate the effectiveness of this approach across various scenarios.

\section*{Impact Statement}
This article introduces a research project aimed at promoting machine learning in federated healthcare systems. Our work may have various social impacts, but none of them are considered significant enough to be specifically highlighted here.

\nocite{langley00}

\bibliography{example_paper}
\bibliographystyle{icml2026}

\newpage
\appendix
\onecolumn
\section{FeDMRA}
Algorithm.~\ref{2} presents the process in Section.~\ref{client continual learning} where the client side fills sample examples according to the allocated category share and updates the local model.
\begin{algorithm}[h]
    \renewcommand{\algorithmicrequire}{\textbf{Input:}}
	\renewcommand{\algorithmicensure}{\textbf{Output:}}
	\caption{FeDMRA: Sample Selection of Example Set}
    \label{2}
    \begin{algorithmic}[1] 
        \REQUIRE  distributed datasets for $t$ tasks: $\{D^1,D^2,...D^t\}$, number of clients: $C$, global model parameters: $w_{g,t-1}$, partitioned storage size: $m_{c,t-1}^y$, initialize the information model parameters: $v_c$, the number of iterations of the client information model: $e$; 
	    \ENSURE local model $w_{c,t}$ and class distribution; 
        \FORALL {$t=1,2,\cdots,T$}
            \STATE  Receive the global model parameters $w_{g,t-1}$ and the allocated storage size $m_{c,t-1}^y$;
               \FOR{$c=1,2,\cdots,C$}
               \STATE $v_{c,e}^{t}\leftarrow$ Take $\bar{D}{_c^{t-1}}$ as input of Eq.(9);
              \STATE $\bar{G}_{x_i} \leftarrow$ The average gradient norm of the sample over $e$ iterations;
              \STATE $D_{c,cache}^{t} \leftarrow$ Sample old samples by sorting them in descending order of the gradient by Eq.(10);
              \STATE $w_{c,t} \leftarrow$ Update the local model with $\bar{D}{_c^{t}}$ by Eq.(13).
               \ENDFOR
        \ENDFOR
    \end{algorithmic}
\end{algorithm}

\section{Datasets and Settings}
\label{app:dataset and task}
\textbf{Datasets.} We used three white blood cell datasets. Among them, Matek-19 contains single white blood cell samples from 100 patients with acute myeloid leukemia and 100 non-leukemia patients. Acevedo-20 was provided by the core laboratory of a hospital in Barcelona and has more than 14,000 images. \cite{dataset} is a high-resolution data set that contains 16,027 annotated normal and pathological white blood cell samples from 78 patients, and we use the alternative name HLwbc hereinafter. 

\textbf{Settings.} We constructed multiple continual learning scenarios based on three datasets, including FCIL, FDIL, and FCDIL. The Tab.~\ref{tab:setting} shows the class settings for each task in each scenario.
\begin{table}[h]
\caption{Task settings of FCIL, FDIL, and FCDIL}
\centering
\begin{tabular}{ccc} 
        \toprule[1.2pt]
        {Task}& \makecell[c]{Class Number per Task}\\ 
       \midrule
        FCIL-M& {4, 3, 3, 3}\\ 
      
        FCIL-A& {4, 3, 3}\\ 
        
        FCIL-H & {4, 2, 2}\\
       
        FDIL & {13, 10, 8}\\ 
   
        FCDIL & {4, 3, 3, 3, 4, 3, 3, 4, 4} \\
        \bottomrule[1.2pt]
    \end{tabular}
\label{tab:setting}
\end{table}

      
        
       
   

\textbf{Baselines.} We compared with the following methods. Among them,  
1) \textbf{FL+iCaRL\cite{icarl}:} An adaptation of iCaRL to federated learning framework, where it computes class prototypes to select and replay exemplars for knowledge preservation, while utilizing the standard FedAvg strategy for model aggregation.
2) \textbf{FL+UACL\cite{UACL}:} A continual learning algorithm for medical image analysis that addresses catastrophic forgetting through a novel exemplar selection strategy for replay. It is also adapted to a federated learning setting by employing the standard FedAvg aggregation method.
3) \textbf{MFCL\cite{MFCL}:} A generative model is trained on the server and then shared with the clients. This model is used to sample synthetic examples of past data, which are incorporated into the training process for new tasks.
4) \textbf{DDDR\cite{DDDR}:} A Latent Diffusion Model is utilized to acquire embeddings from old sample classes. These embeddings are then used to generate pseudo-samples for replay-augmented training.
5) \textbf{ReFed\cite{refed}:} An efficient exemplar selection method designed to address catastrophic forgetting under heterogeneous data distributions.
6) \textbf{UP:} The optimal performance upper bound, a scenario where all historical data from every task is fully accessible. However, this setting serves only as a theoretical reference, as it is both difficult to achieve and impractical in real-world applications. 

\section{Example Memory.}
 Tab.~\ref{tab:cache size} compares the memory allocation strategies for data replay across different approaches. Specifically, UACL, iCaRL, and Re-Fed directly store raw data at the client level, allowing each client to replay 240 historical samples per task, with each class containing 240/$|Y_c^t|$ samples. In contrast, MFCL and DDDR maintain their original replay mechanisms: MFCL incorporates pseudo-samples in each training batch, while DDDR generates pseudo-samples per class. Our proposed method dynamically adjusts the memory allocation as tasks progress. To ensure fair comparison under equivalent global memory constraints, UACL, iCaRL, and Re-Fed strictly adhere to the fixed memory budget $\mathcal{M}$, whereas MFCL and DDDR demonstrate significantly higher total replay quantities that substantially exceed $\mathcal{M}$. 
\begin{table*}[hp]
    \caption{Report the cache size of example-set for each method. Taking FCIL-M as an example.}
    \centering
    \begin{tabular*}{\textwidth}{cp{1.3cm}cccc}
         \toprule[1.2pt]
      {\makebox[0.1\textwidth][c]{\textbf{Method}}} & \multicolumn{4}{c}{\textbf{Memory Size of Each Task}}\\
       \midrule
        iCaRL+FL\textsubscript{(Per-Client)}& 240$\times$5 &240$\times$6 &240$\times$6 &240$\times$7 \\ 
    
        \rule{0pt}{14pt} 
        UACL+FL\textsubscript{(Per-Client)} & 240$\times$5 &240$\times$6 &240$\times$6 &240$\times$7  \\ 
    
        \rule{0pt}{14pt} 
        MFCL\textsubscript{(Per-Batch)} & 128$\times$BS& 128$\times$BS& 128$\times$BS & 128$\times$BS\\
       
        \rule{0pt}{14pt} 
        DDDR\textsubscript{(Per-Class)} & 240$\times$ $Y^t$ &240$\times$ $Y^t$ &240$\times$ $Y^t$ &240$\times$ $Y^t$ \\
       
        \rule{0pt}{14pt} 
        ReFed\textsubscript{(Per-Client)} & 240$\times$5 &240$\times$6 &240$\times$6 &240$\times$7\\ 
        \cdashline{1-5} 
        \rule{0pt}{14pt} 
        FeDMRA\textsubscript{(Per-Client)} & 240$\times$5  & \makecell{322,322,190,209,152,240} & \makecell{311,400,47,309,132,190} &\makecell {346,61,367,122,370,134,240} \\ 
        \bottomrule[1.2pt]
    \end{tabular*}
    \begin{tablenotes}
    \item[] * BS - BatchSize
    \end{tablenotes}
    \label{tab:cache size}
\end{table*}
\section{Computational and Communication Complexity}
We analyze the method complexity and communication complexity from the dimensions of time and space. Assume that the number of model parameters updated by the client is $P$ and the computational complexity per sample is $d$.\\
\textbf{Communication complexity}. The model parameters and category distribution information uploaded by each client in each round:
\begin{equation}
    O\left(R \cdot \left( |C| \cdot P+2(|Y|) \right)\right)
\end{equation}
\textbf{Computational complexity}.
When a new task sequence arrives, the client updates the local information model based on the data from the previous task. Subsequently, it trains the model with the new data. Therefore, the computational complexity of the client is as follows:
\begin{equation}
\begin{aligned}
        &O\left(R\cdot e \cdot \left(\sum_{c} \bar{D_c^t} \cdot d + \sum_{c} \bar{D}{_c^{t-1}} \cdot d\right)\right) \\& \approx O(2R\cdot e \sum_c D_c\cdot d)
\end{aligned}
\end{equation}
The computational complexity when the server dynamically allocates example set space and fuses model parameters is:
\begin{equation}
   O(R\cdot (|C|+1) \cdot P)+O(|Y|)
\end{equation}
Merge Eq. (15) and Eq. (16), the total computational complexity of FeDMRA is:
\begin{equation}
O\left(R\left((2|C| + 1)\cdot P+2|Y|+2e\sum_{c}({D_c}\cdot d)\right)+|Y|\right)
\end{equation}

\section{Discussion} 
We address the issue of exacerbated forgetting in non-IID  data under complex tasks. While conventional data replay schemes raise privacy concerns, we would like to clarify that FeDMRA is not limited to the form of data replay but focuses on allocating storage space for data. In addition, our experiment focuses on the dataset of blood cell disease classification and constructs a rich task scenario. However, relying on a single data set may limit the generality of model for data from other sources. In fact, it is difficult to collect medical datasets from multiple domains.


\end{document}